\documentclass[sigconf,nonacm]{acmart}
\usepackage[utf8]{inputenc}
\usepackage{textcomp}
\usepackage{newunicodechar}
\newunicodechar{≥}{\ensuremath{\geq}}

\usepackage{tikz}
\usepackage{subcaption}

\AtBeginDocument{%
  }

\begin{document}

\title{Actionable Advice from Reviews via Mixture of LoRA Experts: A Two‑LLM Pipeline for Issue Extraction and Business Recommendations}

\author{Kartikey Singh Bhandari}

\affiliation{%
  \institution{BITS Pilani,Pilani Campus}
  \city{Pilani}
  \country{India}}
\email{p20241006@pilani.bits-pilani.ac.in}
\author{Manav Ganesh}

\affiliation{%
  \institution{BITS Pilani,Pilani Campus}
  \city{Pilani}
  \country{India}}
\email{f20211069@pilani.bits-pilani.ac.in}
\author{Yashwant Viswanathan}
\affiliation{%
  \institution{BITS Pilani, Pilani Campus}
  \city{Pilani}
  \country{India}}
\email{f20220599@pilani.bits-pilani.ac.in}
\author{Archit Agrawal}
\affiliation{%
  \institution{BITS Pilani, Pilani Campus}
  \city{Pilani}
  \country{India}
}
\email{f20191048@pilani.bits-pilani.ac.in}
\author{Dhruv Kumar}
\affiliation{%
  \institution{BITS Pilani,Pilani Campus}
  \city{Pilani}
  \country{India}}
  \email{dhruv.kumar@pilani.bits-pilani.ac.in}
\author{Pratik Narang}
\affiliation{%
  \institution{BITS Pilani,Pilani Campus}
  \city{Pilani}
  \country{India}}
  \email{pratik.narang@pilani.bits-pilani.ac.in}






\begin{abstract}
Customer reviews contain detailed, domain-specific signals about service failures and user expectations, but converting this unstructured feedback into actionable business decisions remains difficult. We study review-to-action generation: producing concrete, implementable recommendations grounded in review text. We propose a modular two-LLM framework in which an \emph{Issue} model extracts salient issues and assigns coarse themes, and an \emph{Advice} model generates targeted operational fixes conditioned on the extracted issue representation. To enable specialization without expensive full fine-tuning, we adapt the \emph{Advice} model using a mixture-of-LoRA experts strategy: multiple low-rank adapters are trained and a lightweight gating mechanism performs token-level expert mixing at inference, combining complementary expertise across issue types. We construct synthetic review–issue–advice triples from Yelp reviews (airlines and restaurants) to supervise training, and evaluate recommendations using an eight-dimension operational rubric spanning actionability, specificity, feasibility, expected impact, novelty, non-redundancy, bias, and clarity. Across both domains, our approach consistently outperforms prompting-only and single-adapter baselines, yielding higher actionability and specificity while retaining favorable efficiency–quality trade-offs.
\end{abstract}

\maketitle

\section{Introduction}

Customer reviews are a rich yet underused source of operational insight. While recommender and ranking systems excel with structured behavioral signals (e.g., ratings, clicks, purchases), such signals often under-represent the causal explanations and fine-grained failure modes that appear explicitly in free-form feedback \cite{hasan2025based,mcauley2013hidden}. Consequently, a substantial body of work has focused on \emph{descriptive} review understanding which includes sentiment analysis and opinion mining \cite{pang2009opinion,liu2022sentiment}, aspect-based sentiment analysis \cite{pontiki2016semeval}, and review summarization \cite{angelidis2018summarizing}. More recently, large language models (LLMs) have demonstrated strong capabilities for interpreting nuanced text and generating fluent, instruction-following outputs \cite{brown2020language,ouyang2022training}. However, despite progress in modeling and summarization, review mining rarely produces \emph{prescriptive} outputs: concrete, implementable recommendations that businesses can act on. Bridging this gap is important for web-facing platforms, where feedback streams are heterogeneous and noisy, and the value of analytics depends on whether the output supports real operational decisions.

This paper studies \emph{review-to-action generation}: given raw reviews, generate specific business recommendations that are actionable, non-generic, and grounded in the expressed issues. This task is challenging because high-quality advice must satisfy multiple constraints simultaneously (actionability, specificity, feasibility, impact, clarity), and naive prompting can yield generic or overly verbose suggestions that fail to translate into operational change. From a modeling perspective, the task also stresses \emph{domain and issue diversity}: even within a single platform, complaints vary across industries (e.g., airlines vs.\ restaurants) and across issue types (e.g., delays, staff behavior, cleanliness), motivating methods that can specialize without prohibitive training and deployment costs.

We approach this challenge with a modular, parameter-efficient framework that targets advice quality directly. Our core idea is to combine (i) a structured decomposition of the task into \emph{issue extraction} and \emph{recommendation generation}, with (ii) adapter-based specialization via a \emph{mixture-of-LoRA experts} mechanism. Parameter-efficient fine-tuning (PEFT) methods such as LoRA \cite{hu2022lora} enable adapting large models with small trainable components and modest computational overhead \cite{peft}. However, a single adapter can be brittle under diverse issue distributions and cross-domain settings. Inspired by mixture-of-adapter and mixture-of-experts routing \cite{shazeer2017outrageously,fedus2022switch}, recent work on mixture-of-LoRA experts shows that combining multiple lightweight adapters through gating can improve generalization and robustness \cite{wu2024mixture,buehler2024x}. We adapt these ideas to prescriptive advice generation, enabling token-level expert mixing over frozen adapters so the model can conditionally specialize across issue types without full fine-tuning.

Empirically, our results show consistent gains in \emph{actionability} and \emph{specificity}, with favorable efficiency-quality trade-offs relative to prompting-only and single-adapter baselines. To measure advice quality beyond a single scalar score, we also introduce a multidimensional operational rubric that evaluates recommendations along criteria that matter for execution. Our contributions are:

 \begin{enumerate}
    \item \textbf{Task \& evaluation.} We formalize review to action generation and propose a multidimensional, operational rubric for advice quality.
    \item \textbf{Method.} We adapt mixture-of-LoRA experts technique \cite{wu2024mixture, buehler2024x} to prescriptive advice generation, enabling token-level expert mixing over frozen adapters for specialization without full fine-tuning.
    \item \textbf{System.} We implement a two-LLM, multi-agent based pipeline which extracts issues and proposes actionable fixes.

\end{enumerate}

\subsubsection*{Research Questions}
\begin{enumerate}
    \item RQ1 — Method: Does proposed model improve actionability and specificity relative to single-LoRA models under comparable training budgets?
    \item RQ2 — Generalization: Can a model trained on synthetic airline and restaurants issues produce high-quality advice on other domain without re-training?
\end{enumerate}

\section{Proposed Methodology}
We propose a two-LLM agent framework, as shown in Fig. 1, to process customer reviews and generate business suggestions. The issue-LLM identifies and extracts key information. This LLM condenses unstructured reviews by extracting the issues and themes from those reviews. We pass these issue-theme pairs as inputs to the advice-LLM, which is responsible for suggesting actionable business fixes relevant to the customer reviews. This two-LLM agentic framework ensures that the system is modular and that the outputs of each LLM are more focused on the task assigned to it. By dividing a complex task and assigning it to multiple LLMs, we ensure that performance is improved and output quality is maintained. We then use Mixture of LoRA Experts to fine-tune our advice LLM on synthetic data, further improving the quality of our results.
\begin{figure}[htbp]
\centering
\includegraphics[width=0.45\textwidth]{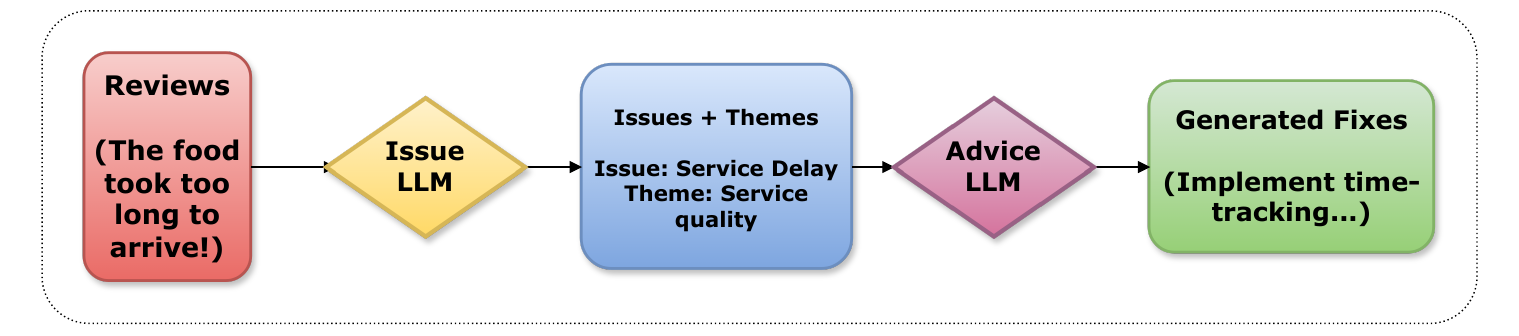}
\caption{Overview of the proposed multi-stage LLM pipeline. User reviews are processed by an Issue LLM to extract specific issues and their corresponding themes, which are then passed to an Advice LLM that generates targeted, actionable fixes.}
\label{fig:Workflow}
\end{figure}

\subsection{Mathematical Formulation}
The main innovation of our advice-LLM is the use of Mixture of LoRA Experts \cite{wu2024mixture,buehler2024x} to finetune this LLM. We first train LoRA adapters for each business industry. These experts are then integrated into the base model $R_{\theta}$ using a gating network \textit{G}, which dynamically selects the most suitable experts by combining the outputs of the most relevant adapters for each given input.

Our architecture consists of a pre-trained Advice-LLM $R_{\theta}$, parameterized by a set of weights $\Theta$, and we denote the parameters of a particular transformer block as $\theta \in \Theta$. We also use an Issue-LLM $I_\phi$ that first process the reviews and then outputs the relevant issues found in the customer reviews, and their themes. The Issue-LLM $I_\phi$ parameterized by weights $\phi \in \Phi$. The outputs of $I_\phi$ are then used as inputs for $R_{\theta}$.

\begin{equation}
    y_{reviews} = R_{\theta}(I_{\phi}(x_{reviews}))
\end{equation}

Where $y_{reviews}$ represents the recommendations generated by the $R_{\theta}$ and $x_{reviews}$ represents the customer review i.e. the input to $I_{\phi}$.

The LoRA adapters, denoted by $\{\Delta\theta_1, \dots, \Delta\theta_N\}$ have been trained on a dataset of customer reviews from different industries. For a given input $\boldsymbol{x} \in \mathbb{R}^{L \times d}$, the output of the model block $\theta$ is represented as $\boldsymbol{F}_{\theta} \in \mathbb{R}^{L \times d}$ and the output of each LoRA is represented as $\boldsymbol{E}_{\Delta\theta_i}(\boldsymbol{x}) \in \mathbb{R}^{L \times d}$.

Mixture of LoRA Experts applies a learnable gating function \textit{G} that determines the weights for each expert based on the input. For each input token \textit{t}, the router computes a set of weights $\{w\}_{i=0}^N$, where w is computed as:

\begin{equation}
    w_i = \frac{\exp(W_g x)_i}{\sum_{j=1}^{N} \exp(W_gx)_j} ,
\end{equation}
Where $W_g$  is a learnable parameter and x represents the input for which the gating weights are being calculated. This weighted expert output along with the base model's output comprises the Mixture of LoRA Experts output represented by $O$.

\begin{equation}
O = F_{\theta} + \sum_{i=1}^{N} w_i(E_{\Delta \theta_i}(x)) 
\label{eq:mole_output}
\end{equation}

\begin{figure}[htbp]
\centering
\includegraphics[width=0.5\textwidth]{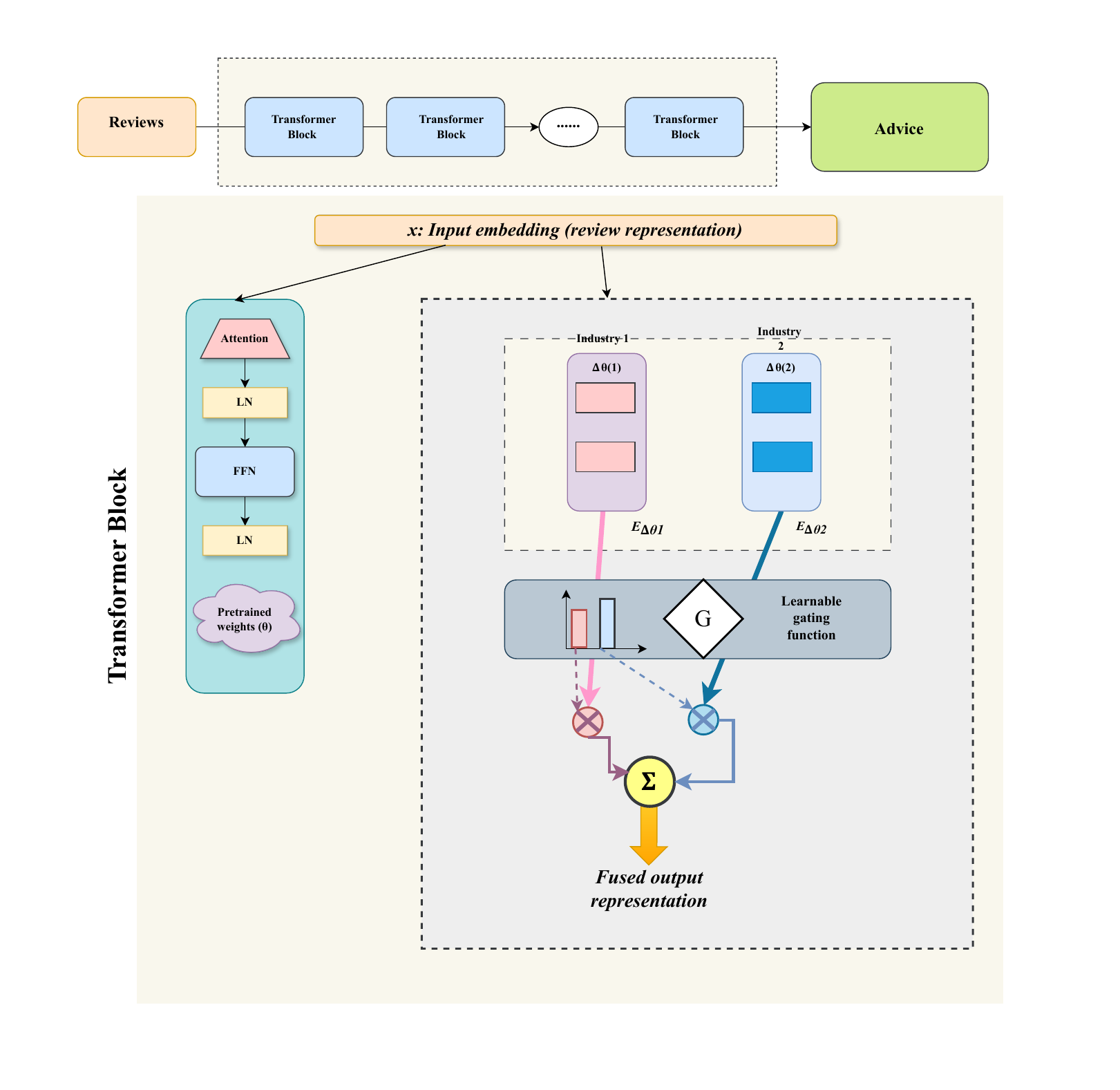}
\caption{Architecture of the proposed gated multi-industry transformer for review to fix generation.
User reviews are encoded into input embeddings that propagate through a pretrained transformer stack. Each transformer block augments the shared parameters ($\theta$) with industry-specific offsets $\Delta\theta(i)$, producing specialised representations $E_{\Delta \theta_i}$. A learnable gating function G adaptively combines these industry-conditioned outputs into a single fused representation, which is then passed through the remaining transformer layers to generate context- and domain-aligned fixes.}
\label{fig:Workflow2}
\end{figure}
\subsection{Synthetic Data Generation}
Synthetic data has been used to train our network. We utilised publicly available Yelp\footnote{https://business.yelp.com/data/resources/open-dataset/} reviews and categorised them into positive and negative reviews. We then used the negative reviews as inputs for our multi-agentic workflow and generated a synthetic dataset with reviews, themes and fixes. GPT-OSS 120B\footnote{https://huggingface.co/openai/gpt-oss-120b} is used for both issue and advice LLMs to ensure that a larger, more capable model can generate a rich and high-quality dataset, which is essential for finetuning purposes. 

Once the synthetic data is ready, we use a fine-tuned Llama3.1-8B-instruct as the advice LLM. This methodology employs a form of knowledge-distillation, where a smaller model learns to mimic the performance of a larger model using only a fraction of the compute.
 
\subsection{Training}
The base model used is the 4-bit quantized Llama 3.1-8B-instruct\footnote{https://huggingface.co/meta-llama/Llama-3.1-8B-Instruct}. The LoRA adapters are trained on customer review data for airlines and restaurants separately, and these adapters are then combined and attached to the base model using mixture of LoRA experts. We then freeze the base model and retrain the LoRA adapters and the gating network on the combined synthetic data set to ensure that the outputs of the advice LLM are of high quality, and that the model is able to decide which expert to consult based on the input token.

\section{Evaluation}
We evaluate each recommendation on eight 0–100 rubric dimensions. The dimensions are Actionability, Specificity, Feasibility, Expected Impact, Novelty, Non-redundancy, Bias and Reading Clarity. Each dimension is rated on a 5-point Likert scale and linearly rescaled to 0–100 via
\[
s_{\text{scaled}} \;=\; 100 \times \frac{(x-1)}{4}\,, \quad x \in \{1,2,3,4,5\} . \tag{5}\] and the overall score is the simple arithmetic mean of the eight scaled dimensions.
\section{Results}
\begin{figure}[htbp]
\centering
\includegraphics[width=0.45\textwidth]{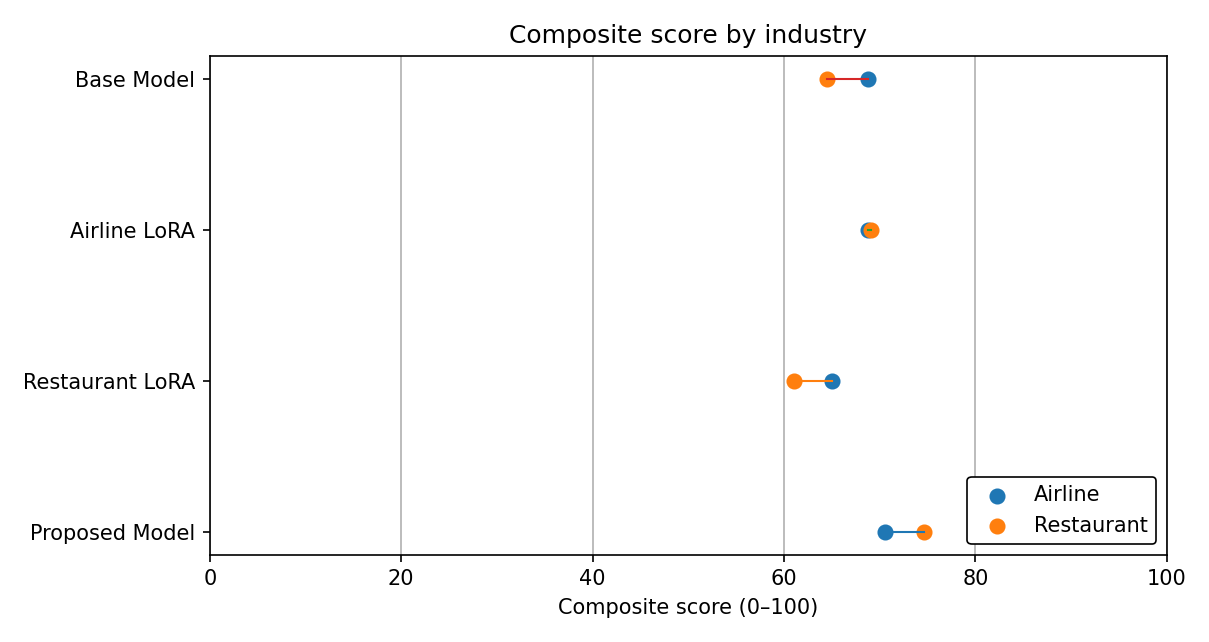}
\caption{Composite scores (0–100) for each model in the Airline and Restaurant industries. Points denote the industry-specific composite (mean across eight dimensions), and horizontal connectors link the same model across industries to highlight cross-domain differences.}
\label{fig:Workflow}
\end{figure}

\begin{table*}[t]
\centering
\small
\setlength{\tabcolsep}{3.5pt}
\begin{tabular}{lrrrrrrrr}
\toprule
 & \multicolumn{4}{c}{Airline} & \multicolumn{4}{c}{Restaurant} \\
\cmidrule(lr){2-5}\cmidrule(lr){6-9}
Dimension & Base & Air LoRA & Res LoRA & Proposed & Base & Air LoRA & Res LoRA & Proposed \\
\midrule
actionability   & 70.0 & 70.0 & 65.0 & 75.0 & 68.8 & 75.0 & 72.2 & 75.0 \\
specificity     & 60.0 & 75.0 & 80.0 & 70.0 & 50.0 & 63.9 & 55.6 & 78.6 \\
feasibility     & 70.0 & 30.0 & 30.0 & 30.0 & 50.0 & 36.1 & 30.6 & 46.4 \\
expected impact & 70.0 & 60.0 & 60.0 & 65.0 & 65.6 & 58.3 & 50.0 & 67.9 \\
novelty         & 30.0 & 65.0 & 45.0 & 65.0 & 46.9 & 63.9 & 50.0 & 57.1 \\
non redundancy  & 70.0 & 75.0 & 65.0 & 75.0 & 71.9 & 69.4 & 63.9 & 82.1 \\
bias            & 100.0 & 100.0 & 95.0 & 100.0 & 100.0 & 97.2 & 88.9 & 96.4 \\
reading clarity & 80.0 & 75.0 & 80.0 & 85.0 & 62.5 & 88.9 & 77.8 & 92.9 \\
\midrule
Composite (avg.) & 68.8 & 68.8 & 65.0 & 70.6 & 64.5 & 69.1 & 61.1 & 74.6 \\
\bottomrule
\end{tabular}
\caption{Dimension scores and composite (mean across the 8 dimensions) for Airline and Restaurant industries.}
\label{tab:dimension_composite_scores}
\end{table*}

\subsection{Airline}

We evaluate four systems on airline reviews: the base model, an airline-specific LoRA, a restaurant LoRA used cross-domain, and the proposed model. Table \ref{tab:dimension_composite_scores} shows that the base model and airline LoRA obtain the same composite score (68.8), the restaurant LoRA performs worse (65.0), and the proposed model performs best (70.6). As shown in Figure Y, all models maintain high bias (≥95) and good reading clarity (≥75). The proposed model achieves the strongest actionability (75), high novelty (65), and the best or tied-best non-redundancy and clarity, indicating more actionable, non-repetitive, and readable advice than the baselines. However, all fine-tuned models sharply reduce feasibility from 70 (base) to around 30, reflecting a shift toward detailed, tool-heavy recommendations that are harder to implement.

\begin{figure}[htbp]
\centering
\includegraphics[width=0.45\textwidth]{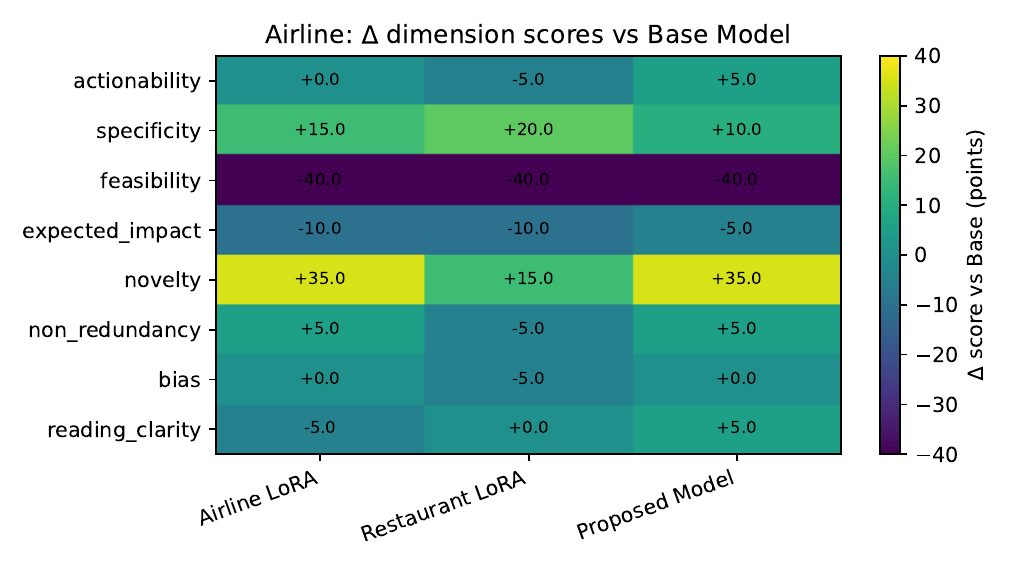}
\caption{Airline industry: per-dimension score changes relative to the Base Model for Airline LoRA, Restaurant LoRA, and the Proposed Model. Each cell reports the signed difference in points (model - base), showing large gains in novelty (up to +35) and specificity (+10 to +20), but a consistent drop in feasibility (-40) across all adapted models; the Proposed Model also improves actionability (+5) and reading clarity (+5).}
\label{fig:Workflow}
\end{figure}


\subsection{Restaurant}
On restaurant reviews, we again compare the base model, an airline-domain LoRA, a restaurant-domain LoRA, and the proposed model. As shown in Table \ref{tab:dimension_composite_scores}, the base model scores 64.5 on the composite metric, the airline LoRA improves this to 69.1, and the restaurant LoRA drops to 61.1. The proposed model achieves 74.6, yielding gains of +10.1 over the base, +5.5 over the airline LoRA, and +13.5 over the restaurant LoRA.

Table \ref{tab:dimension_composite_scores} shows that the proposed model matches or outperforms all baselines on most dimensions, with the highest specificity (78.6), non-redundancy (82.1), and reading clarity (92.9), and the best expected impact (67.9). Actionability is also strong (75.0), comparable to the airline LoRA and higher than the base and restaurant LoRA. As in the airline domain, feasibility remains lower than the base (46.4 vs. 50.0), suggesting that more detailed recommendations are judged harder to implement, while all models maintain high bias scores (≥88.9), indicating that adapter training does not degrade safety.
\begin{figure}[htbp]
\centering
\includegraphics[width=0.45\textwidth]{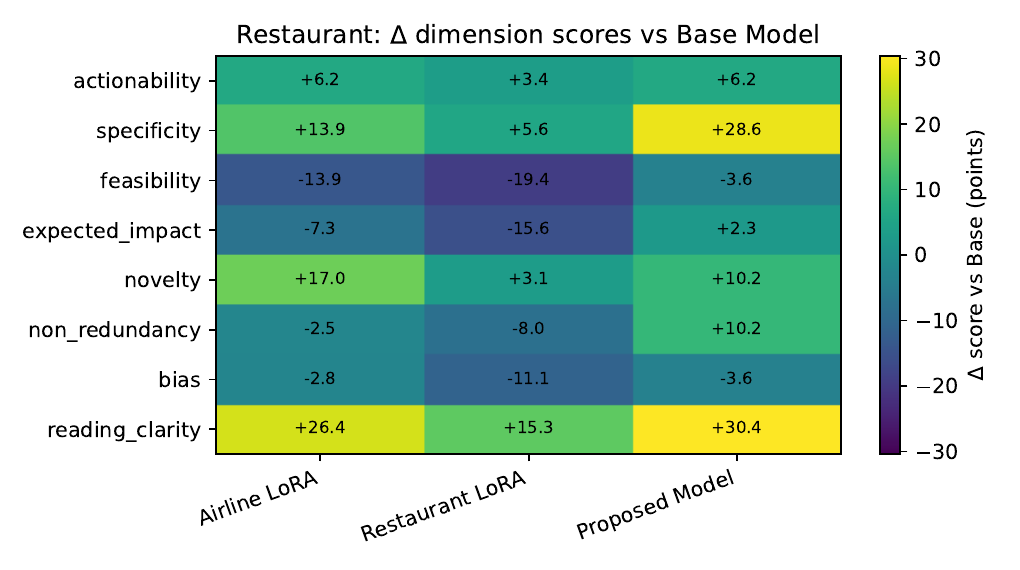}
\caption{Restaurant industry: per-dimension score changes relative to the Base Model for Airline LoRA, Restaurant LoRA, and the Proposed Model. Each cell reports the signed difference in points (model - base), highlighting that the Proposed Model delivers the largest gains in specificity (+28.6) and reading clarity (+30.4), while feasibility decreases across all models (-3.6 to -19.4).}
\label{fig:Workflow}
\end{figure}

\section{Discussion}
\subsection{Effects of Adapter Design in the Airline Domain}

For RQ1, the airline results show that simply adding a domain-specific LoRA on top of a strong instruction-tuned base model does not necessarily improve the overall quality of business recommendations: the airline LoRA matches, but does not exceed, the base model on the composite metric and introduces a strong trade-off between specificity and feasibility. In contrast, the proposed model yields a consistent, albeit moderate, improvement in the composite score and dominates the baselines on actionability, novelty, non-redundancy, and reading clarity. At the same time, all fine-tuned variants suffer large drops in feasibility, likely because the synthetic training data encourages very detailed, tool-stacked interventions that the judge deems harder to implement. This suggests that mixture-of-adapter architectures can better harness the benefits of domain specialization while preserving overall response quality, but future work should explicitly regularize for implementability to avoid over-engineering in the generated advice.

\subsection{Cross-Domain vs. In-Domain Adaptation}

For restaurant reviews, the results provide a nuanced view of RQ1 and RQ2. A single in-domain adapter (restaurant LoRA) does not yield the best performance and in fact underperforms both the base model and the cross-domain airline LoRA on the composite metric, likely because it overfits to synthetic training patterns that emphasize enterprise service scenarios rather than front-of-house restaurant workflows. By contrast, the airline LoRA transfers surprisingly well to restaurants, improving over the base model on actionability, novelty, and clarity, which suggests that some operational patterns—such as managing queues, coordinating staff, and handling complaints—generalize across service industries. The proposed model delivers the strongest gains by blending multiple adapters, capturing this cross-domain structure while also regularizing against the domain drift seen in the restaurant-only adapter. However, the persistent drop in feasibility relative to the base model indicates that future work should explicitly encourage simpler, easier-to-execute interventions, for example by penalizing excessively tool-stacked plans or incorporating human-in-the-loop feasibility judgments during training.

\section{Conclusion}
We presented a two-stage LLM pipeline that turns free-form reviews into structured business recommendations. The system combines an issue extraction model with a recommendation model built from a mixture of lightweight adapters on a quantized base LLM, trained via synthetic review–issue–advice triples and evaluated with an LLM-as-a-judge rubric over eight quality dimensions.

On airline and restaurant reviews, the proposed model consistently outperforms a strong base model and single-adapter baselines, especially in actionability, specificity, non-redundancy, and reading clarity, while maintaining high bias scores. At the same time, all fine-tuned variants expose a tension between detailed, tool-heavy suggestions and perceived feasibility, and we observe that some adapters transfer across service domains more robustly than purely in-domain ones. These results highlight mixture of adapters architectures as a promising, deployment-friendly approach for extracting actionable insight from web reviews. Future work will incorporate feasibility and cost-aware training signals and extend the pipeline to additional verticals and human in the loop evaluation.
\bibliographystyle{ACM-Reference-Format}
\bibliography{sample-base}
\end{document}